\documentclass[letterpaper, 10 pt, conference]{ieeeconf}  

\IEEEoverridecommandlockouts                              
\usepackage{graphics} 
\usepackage{epsfig} 
\usepackage[noadjust]{cite}
\usepackage{amsmath}
\usepackage{graphicx}
\usepackage{amssymb}
\usepackage{colortbl}
\usepackage{arydshln}
\usepackage{stfloats}
\usepackage{longtable}

\usepackage[vlined,boxed,ruled,linesnumbered]{algorithm2e}
\SetKwProg{Fn}{Function}{}{}
\usepackage{algpseudocode}

\newtheorem{problem}{Problem}

\newtheorem{assumption}{Assumption}

\newtheorem{lemma}{Lemma}
\newtheorem{remark}{Remark}
\newtheorem{proposition}{Proposition}

\title{\LARGE \bf
Robust Optimization Framework for Training Shallow Neural Networks Using Reachability Method 
}

\author{Yejiang Yang and Weiming Xiang,~\IEEEmembership{Senior Member, IEEE} 
\thanks{Yejiang Yang is with Department of Electrical Engineering, Southwest Jiaotong University, China
        {\tt\small yangyejiang0316@163.com}}%
\thanks{Weiming Xiang is with School of Computer and Cyber Sciences, Augusta University, Augusta GA 30912, USA
        {\tt\small wxiang@augusta.edu}}%
}

\begin{document}

\maketitle
\thispagestyle{empty}
\pagestyle{empty}

\begin{abstract}
In this paper, a robust optimization framework is developed to train shallow neural networks based on reachability analysis of neural networks. To characterize noises of input data, the input training data is disturbed in the description of interval sets. Interval-based reachability analysis is then performed for the hidden layer. With the reachability analysis results, a robust optimization training method is developed in the framework of robust least-square problems. Then, the developed robust least-square problem is relaxed to a semi-definite programming problem. It has been shown that the developed robust learning method can provide better robustness against perturbations at the price of loss of training accuracy to some extent. At last, the proposed method is evaluated on a robot arm model learning example.    
\end{abstract}

\section{Introduction}
Neural networks are extensively used in machine learning systems for their effectiveness in controlling complex systems in a variety of research activities such as focusing on the promise of neural networks in controlling nonlinear systems in \cite{hunt1992neural}, providing a neural network control method for a general type of nonlinear system in \cite{ge1999adaptive}, controlling industrial processes using adaptive neural network in \cite{wang2015combined}, investigating the problem of sampled data stabilization for neural network control systems with a guaranteed cost in \cite{wu2014exponential}, controlling autonomous vehicles using deep neural networks in \cite{tian2018deeptest}, etc. However, small perturbations in the input can significantly distort the output of the neural network. 

The robustness and reliability of neural networks have received particular attention in the machine learning community. For instance, adding Lyapunov constraints in the training neural networks to enhance stabilization of learning nonlinear system in \cite{neumann2013neural}, introducing performance-driven BP learning for edge decision-making in \cite{he2020learnability}, studying the adversarial robustness of neural networks using robust optimization \cite{madry2017towards}, adversarial neural pruning and suppressing latent vulnerability by proposing a Bayesian framework in \cite{madaan2020adversarial}, studying the method to learn deep ReLU-based classifiers that are provably robust against norm-bounded adversarial perturbations in \cite{wong2018provable}, identifying a trade-off between robustness and accuracy in the design of defenses against adversarial examples in \cite{zhang2019theoretically}, an intelligent bearing fault diagnosis method based on stacked inverted residual convolution neural networks  is proposed in \cite{yao2020lightweight}, etc. Yet by adding adversarial examples in the training data may not improve the worst-case performance of neural networks directly, and existing robust optimization frameworks for training neural networks are focusing on training classifiers. There's an increasing demand for neural networks with a certain degree of immunity to perturbations and the method for guaranteed approximation error estimation between the system and its neural network approximation in safety-critical scenarios. The key to training robust neural networks is to mitigate perturbations. Based on the methods proposed in \cite{xiang2017reachable,xiang2018output,xiang2018reachable,tran2019star,xiang2018reachability}, the output reachability analysis of neural networks can be performed in presence of perturbations in input data .

According to the Universal Approximation Theorem, a shallow neural network could be able to solve any nonlinear approximation problems \cite{huang2006extreme}. In this work, a shallow neural network is considered for saving computing resources with its simple structure and the less conservative output of which can be obtained using reachability methods. This paper aims to train a robust neural network by mitigating the effects caused by perturbations using reachable set methods. Specifically, we form the output reachable set from the input disturbed data described by the interval-based reachability method. After forming the reachable set of the hidden layer, the training of shallow neural network is performed by solving robust least-square problems as an extension of the Extreme Learning Machine (ELM) method proposed in \cite{huang2006extreme}. The neural network trained by the proposed robust optimization framework is then compared with ELM in terms of reachable set estimation and the mean square error in presence of perturbations in input data, as shown in a robot modeling example. The results indicate that the robustness using the proposed method has been improved and the trade-off between accuracy and robustness is tolerable when it comes to robust data-driven modeling problems.
 
The rest of the paper is organized as follows: preliminaries and problem formulation are given in Section II. The main result, robust optimization training for shallow neural networks using reachability method is presented in Section III. In Section IV, a robot arm modeling example is provided to evaluate our method. Conclusions are given in Section V.

\emph{Notations:}  $\mathbb{N}$ denotes the set of natural numbers, $\mathbb{R}$ represents the field of real numbers, and $\mathbb{R}^{n}$ is the vector space of all $n$-tuples
of real numbers, $\mathbb{R}^{n \times n}$ is the space of $n\times n$ matrices with real entries. Given a matrix $X \in \mathbb{R}^{n \times n}$, the notation $X \succ 0$ means $X$ is real symmetric and positive definite. Given a matrix $A \triangleq [a_{i,j}]\in \mathbb{R}^{n \times m}$, $\mathrm{vec(A)} = [a_{1,1},\cdots,a_{n,m}]^{\top}$ is the vectorization of $A$. $\left\|\cdot\right\|_2$ and $\left\|\cdot\right\|_\infty$ stand for Euclidean norm and infinity norm, respectively. In addition, in symmetric block matrices, we use * as an ellipsis for the terms that are induced by symmetry.

\section{Preliminaries and Problem formulation}
\subsection{Shallow Feedforward Neural Networks}
A neural network consists of  hidden layers with information flow from input layer to output layer. Each layer consists of processing neurons which respond to the weighted input receiving from other neurons in the form of:
\begin{align}\label{single layer}
    x_i=\phi(\sum\nolimits^n_{j=1}w_{i,j}x_{j}+b_i)
\end{align}
where $x_j\in\mathbb{R}$ is the $j$th input of the $i$th neuron,  $n\in\mathbb{N}$ is the number of inputs for the $i$th neuron, $x_i\in\mathbb{R}$ is the output of the $i$th neuron, $w_{i,j}\in\mathbb{R}$ is the weight from $j$th neuron to $i$th neuron, $b_i\in\mathbb{R}$ is the bias of the $i$th neuron, and  $\phi:\mathbb{R}\rightarrow\mathbb{R}$ is the activation function.

The feedforward neural network considered in this paper is a class of neural networks with no cycles and loops. Each layer is connected with nearby layer by the weight matrix and with a bias vector in the form of
\begin{align}
W^{(k)} &= \begin{bmatrix}
    w^{(k)}_{1,1} & w^{(k)}_{1,2} & \cdots & w^{(k)}_{1,n_{k-1}}
    \\
    w^{(k)}_{2,1} & w^{(k)}_{2,2} & \cdots & w^{(k)}_{2,n_{k-1}}
    \\
    \vdots & \vdots & \ddots & \vdots
    \\
    w^{(k)}_{n_k,1} & w^{(k)}_{n_k,2} & \cdots & w^{(k)}_{n_k,n_{k-1}}
    \end{bmatrix}
    \\
 b^{(k)}&=[b_1^{(k)},\cdots,b_{n_k}^{(k)}]^\mathrm{\top}
\end{align}
in which $n_k\in\mathbb{N}$ denotes the number of neurons in layer $k$, thus the output of $k$th layer $x^{(k)}$ can be described by
\begin{align}\label{nn}
    x^{(k)}=\phi_k(W^{(k)}x^{(k-1)}+b^{(k)}),\ \forall k =1,\cdots,L
\end{align}
where $x^{(k)} \in \mathbb{R}^{n_k}$ is the output vector of layer $k$. 

Specifically, we consider a shallow feedforward neural network with one hidden layer $\phi_1$ and the output layer $\phi_2$, the mapping from the input vector $x^{(0)}$ of input layer to the output vector $x^{(2)}$ can be expressed in the form of 
\begin{align} \label{shallow}
    x^{(2)}=\Phi(x^{(0)})
\end{align}
where $ \Phi(\cdot) = \phi_{2} \circ \phi_{1} $.
\begin{remark}
Compared with deep neural networks with multiple hidden layers, shallow neural networks normally consist of one hidden layer. Shallow neural networks represent a significant class of neural networks in the machine learning field, e.g., Extreme Learning Machine (ELM) \cite{huang2006extreme}, Broad Learning Systems (BLS) \cite{chen2017broad}. Notably, these shallow neural networks are learners with universal approximation and
classification capabilities provided with sufficient neurons and data, e.g., as shown in \cite{huang2006extreme}, \cite{chen2017broad}. 
\end{remark}

\subsection{Reachability of Neural Networks}
In this paper, we introduce the concept of reachable set of neural networks for robust training. Some critical notions and results related to the reachability of neural networks are presented as follows. 

Given interval set $\mathcal{X}^{(k)}=[\underline{x}^{(k)},\overline{x}^{(k)}]$ of $k$th layer, in which $\underline{x}^{(k)},\overline{x}^{(k)}\in\mathbb{R}^{n_k}$ are the lower- and upper-bond of the $k$th layer's output. The interval set $\mathcal{X}^{(k+1)}=[\underline{x}^{(k+1)}, \overline{x}^{(k+1)}]$ for layer $k+1$ of is defined by
\begin{align}\label{reachable set computation}
\mathcal{X}^{(k+1)}=[\phi_k](\mathcal{X}^{(k)}) .
\end{align}

Specifically, the following trivial assumption, which is satisfied by most of the activation functions, is given for the computation of set $\mathcal{X}^{(k+1)}$. 

\begin{assumption}\label{assumption_mono}
Assume that the following inequality for activation function 
\begin{align}\label{mono}
    \phi_k(x_1)\le \phi_k(x_2)
\end{align}
holds for any two scalars $x_1 \le x_2$.
\end{assumption}

With Assumption \ref{assumption_mono}, 
$\underline{x}^{(k+1)}_i$ and $\overline{x}^{(k+1)}_i$ in  $\mathcal{X}^{(k+1)} = [\underline{x}^{(k+1)}, \overline{x}^{(k+1)}]$ can be computed by the following lemma, which is inspired by \cite{xiang2018output,xiang2020reachable}. 
\begin{lemma}\label{lem_outputset}
Given a shallow feedforward neural network (\ref{shallow}) and an input set $\mathcal{X}^{(0)}=[\underline{x}^{(0)},\overline{x}^{(0)}]$, then the output reachable set of $\mathcal{X}^{(1)}$ and $\mathcal{X}^{(2)}$ can be computed by
\begin{align}
\underline{x}^{(k)}_i &= \phi_k(\sum\nolimits_{j=1}^{n_{k}}\underline{p}^{(k)}_{i,j}+b_i)\\
\overline{x}^{(k)}_i  &= \phi_k(\sum\nolimits_{j=1}^{n_{k}}\overline{p}^{(k)}_{i,j}+b_i)
\end{align}
where ${p}^{(k)}_{i,j}$ and $\overline{p}^{(k)}_{i,j}$ are
\begin{align}
    \underline{p}^{(k)}_{i,j}&=\begin{cases}w^{(k)}_{i,j}\underline{x}_j^{(k-1)} & w^{(k)}_{i,j} \ge 0 
    \\
    w^{(k)}_{i,j}\overline{x}_j^{(k-1)} & w^{(k)}_{i,j} < 0
    \end{cases}
    \\
    \overline{p}^{(k)}_{i,j} &=\begin{cases}w^{(k)}_{i,j}\overline{x}_j^{(k-1)} & w^{(k)}_{i,j} \ge 0 
    \\
    w^{(k)}_{i,j}\underline{x}_j^{(k-1)} & w^{(k)}_{i,j} < 0 
    \end{cases} 
\end{align}
where $k = 1,2$. 
\end{lemma}

\begin{proof}
  The proof is given in Appendix A.
\end{proof}
\begin{remark}
Lemma 1 provides a formula to compute the reachable set for both the hidden layer as well as output layer. As indicated in \cite{xiang2020reachable}, the interval arithmetic computation framework might lead to overly conservative over-approximation as the number of hidden layers grows large. However, for shallow neural networks considered in this paper, there is only one hidden layer so that this interval arithmetic computation framework proposed in Lemma \ref{lem_outputset} performs well in practice. 
\end{remark}

\subsection{Problem Formulation}
This paper aims to provide the method of training neural networks to enhance their robustness and reliability using the reachable set. 

Given $N$ arbitrary distinct input-output samples $\{u_i,y_i\}$ with $u_i \in \mathbb{R}^{n_0}$ and $y_i \in \mathbb{R}^{n_2}$, shallow neural network training aims to find weights and biases $W^{(k)}$, $b^{(k)}$, $k = 1,2$ for the following optimization problem of 
\begin{align}
    \min_{W^{(k)}, b^{(k)},k=1,2}\left\|\mathrm{vec}(\Phi(U) - Y) \right\|_2
\end{align}
where $U$ and $Y$ are
\begin{align}  \label{U}
U &= 
    \begin{bmatrix}
    u_1
    \\
    \vdots
    \\
    u_N
    \end{bmatrix}
    =
    \begin{bmatrix}
    u_{1,1} & \cdots & u_{1,n^{(0)}} \\
    \vdots & \ddots & \vdots \\ 
    u_{N,1} & \cdots & u_{N,n^{(0)}} 
    \end{bmatrix}
    \\
    Y &= 
    \begin{bmatrix}
    y_1^{\top}
    \\
    \vdots
    \\
    y_N^{\top}
    \end{bmatrix}
    =
    \begin{bmatrix}
    y_{1,1} & \cdots & y_{1,n^{(2)}} \\
    \vdots & \ddots & \vdots \\ 
    y_{N,1} & \cdots & y_{N,n^{(2)}} 
    \end{bmatrix}. \label{Y}
\end{align}

In this paper, we utilize the ELM proposed in \cite{huang2006extreme} to train shallow neural networks. 
Unlike the most
common understanding that all the parameters of neural networks, i.e., $W^{(k)}$, $b^{(k)}$, $k = 1,2$, need to be adjusted, weights $W^{(1)}$, and the hidden
layer biases $b^{(1)}$ are in fact not necessarily tuned and can actually remain unchanged once random values have been assigned to these parameters in the beginning of training. By further letting $b^{(2)} = 0$ and linear functions as output activation functions as in ELM training, one can rewrite $\Phi(U) = W^{(2)}H(U)$ where
\begin{align} \label{H}
   H(U) = 
   \begin{bmatrix} 
   h(u_1) \\
   \vdots
   \\
   h(u_N) 
   \end{bmatrix}^{\top} =
   \begin{bmatrix} 
h_{1,1}(u_1)  & \cdots & h_{1,n^{(1)}}(u_1) \\ 
\vdots & \ddots & \vdots \\ 
h_{N,1}(u_N)  & \cdots & h_{N,n^{(1)}}(u_N) 
\end{bmatrix}^{\top}
\end{align}
in which $h(u_i) = \phi_1(W^{(1)}u_i+b^{(1)})$. Then, 
the training process is then can be formulated in the form of 
\begin{align}\label{elm_opt}
    \min_{ W^{(2)} }\left\|\mathrm{vec}(W^{(2)}H(U) - Y)\right\|_2 .
\end{align}

To incorporate robustness in the training process in (\ref{elm_opt}), the neural network is expected to be robust to the disturbances injected in the input. Therefore, input data are generalized from points to intervals containing perturbations in the data, i.e., input data $u_i$ are purposefully crafted to $[\underline{u}_i,\overline{u}_i]$ where $\underline{u}_i = u_i -\delta_i$, $\overline{u}_i = u_i +\delta_i$ with $\delta_i > 0$ representing perturbations. The interval data set is denoted by $\mathcal{U}$. Moreover, the trained neural network is expected to be capable of mitigating the changes caused by perturbations as small as possible, thus the target data set is expected to stay the same, i.e., $Y$. 

With the interval data set $\mathcal{U}$, the robust training problem can be stated as follows, which is the main problem to be addressed in this paper.

\begin{problem}
Given $N$ arbitrary distinct input-output samples $\{u_i,y_i\}$ with $u_i \in \mathbb{R}^{n_0}$ and $y_i \in \mathbb{R}^{n_2}$, and also considering perturbations $\delta_i$, how does one compute  weights and biases $W^{(k)}$, $b^{(k)}$, $k = 1,2$ for the following robust optimization problem of 
\begin{align} \label{robust_opt}
    \min_{W^{(k)},b^{(k)},k=1,2}\max_{\delta_i,i = 1,\ldots N}\left\|\mathrm{vec}(\Phi(\mathcal{U}) - Y )\right\|_2
\end{align}
where $Y$ is defined by (\ref{Y}) and $\mathcal{U}$ are
\begin{align} \label{U_interval}
\mathcal{U} = 
    \begin{bmatrix}
    [\underline{u}_1^{\top}, \overline{u}_1^{\top}]
    \\
    \vdots
    \\
       [\underline{u}_N^{\top}, \overline{u}_N^{\top}]
    \end{bmatrix} = 
    \begin{bmatrix}
    [\underline{u}_{1,1},\overline{u}_{1,1}] & \cdots & [\underline{u}_{1,n^{(0)}},\overline{u}_{1,n^{(0)}}] \\
    \vdots & \ddots & \vdots \\ 
     [\underline{u}_{N,1},\overline{u}_{N,1}] & \cdots &  [\underline{u}_{N,n^{(0)}},\overline{u}_{N,n^{(0)}}]
    \end{bmatrix}
\end{align}
in which  $\underline{u}_i = u_i -\delta_i$, $\overline{u}_i = u_i +\delta_i$ with $\delta_i > 0$. 
\end{problem}
\begin{remark}
From (\ref{robust_opt}), the weights and biases in neural networks are optimized to mitigate the negative effects brought in by perturbations in the training process. Moreover, in this paper, we will utilize ELM for shallow neural network training, thus the robust optimization problem (\ref{robust_opt}) can be converted to the following optimization problem
\begin{align}\label{robust_elm_opt}
    \min_{ W^{(2)}}\max_{\delta_i=1,\ldots,N} \left\|\mathrm{vec}(W^{(2)}H(\mathcal{U}) - Y)\right\|_2 
\end{align}
where 
\begin{align} \label{H_interval}
   H(\mathcal{U}) &= 
   \begin{bmatrix} 
   [h]([\underline{u}_1,\overline{u}_1]) \\
   \vdots
   \\
   [h]([\underline{u}_N,\overline{u}_N])
   \end{bmatrix}^{\top}   \nonumber
   \\
  & =
      \begin{bmatrix} 
   [\underline{h}_{1,1},\overline{h}_{1,1}] & \cdots & [\underline{h}_{1,n^{(1)}},\overline{h}_{1,n^{(1)}} ]
   \\
    \vdots & \ddots & \vdots \\ 
    [\underline{h}_{N,1},\overline{h}_{N,1}] & \cdots & [\underline{h}_{N,n^{(1)}},\overline{h}_{N,n^{(1)}} ]  
   \end{bmatrix}^{\top} . 
   \end{align} 

Compared with optimization problem (\ref{elm_opt}) for ELM, the above robust optimization problem (\ref{robust_elm_opt}) can be viewed as an extension for ELM training that involves perturbations in input data.  
\end{remark}

\section{Robust Optimization Training Framework}

In this section, a robust optimization-based training method for shallow neural networks will be presented. As we utilize the ELM training framework as proposed in \cite{huang2006extreme}, the hidden layer weights $W^{(1)}$, and the hidden layer biases $b^{(1)}$ can be randomly assigned. With the random assignment of these parameters and using the reachability results proposed in Lemma \ref{lem_outputset}, $H(\mathcal{U})$ can be obtained. 

\begin{proposition} \label{proposition1}
Given a shallow feedforward neural network (\ref{shallow}) and a disturbed input data set $\mathcal{U}$ described by (\ref{U_interval}), then the interval matrix $H(\mathcal{U})$ in the form of (\ref{H_interval}) can be computed by
\begin{align} \label{propostion1_1}
\underline{h}_{i,j} &= \phi_1(\sum\nolimits_{k=1}^{n^{(1)}}\underline{p}^{(1)}_{i,j,k}+b_j)\\
\overline{h}_{i,j}  &= \phi_1(\sum\nolimits_{k=1}^{n_{1}}\overline{p}^{(1)}_{i,j,k}+b_j) \label{propostion1_2}
\end{align}
where $\underline{p}^{(1)}_{i,j,k}$ and $\overline{p}^{(1)}_{i,j,k}$ are
\begin{align} \label{propostion1_3}
    \underline{p}^{(1)}_{i,j,k}&=\begin{cases}w^{(1)}_{j,k}\underline{u}_{i,j} & w^{(1)}_{j,k} \ge 0 
    \\
    w^{(1)}_{j,k}\overline{u}_{i,j} & w^{(1)}_{j,k} < 0
    \end{cases}
    \\ \label{propostion1_4}
    \overline{p}^{(1)}_{i,j,k} &=\begin{cases}w^{(1)}_{j,k}\overline{u}_{i,j} & w^{(1)}_{j,k} \ge 0 
    \\
    w^{(1)}_{j,k}\underline{u}_{i,j} & w^{(1)}_{j,k} < 0 
    \end{cases} 
\end{align}
and $\phi_1$ is the activation function satisfying Assumption \ref{assumption_mono}.  
\end{proposition}

\begin{proof}
Using the results in Lemma \ref{lem_outputset}, (\ref{propostion1_1})--(\ref{propostion1_4}) can be obtained straightforwardly by letting $k=1$,  $\mathcal{X}^{(0)} = \mathcal{U}$ and $\mathcal{X}^{(1)} =H(\mathcal{U})$. The proof is complete. 
\end{proof}

With the interval matrix $H(\mathcal{U})$, the next critical step in robust training of shallow neural networks is solving robust optimization problem (\ref{robust_elm_opt}) to compute weights $W^{(2)}$ of output layer.
\begin{proposition}\label{proposition2}
Given a shallow feedforward neural network (\ref{shallow}), a disturbed input data set $\mathcal{U}$ described by (\ref{U_interval}) and a target data set $Y$, then the interval matrix $H(\mathcal{U})$ for hidden layer is given in the form of (\ref{H_interval}), there exist scalars $\gamma > 0$, $\lambda_{i,j} > 0$ and matrix $W^{(2)}$  such that $\left\|\mathrm{vec}(W^{(2)}H(\mathcal{U}) - Y)\right\|_2 \le \gamma$ where $\gamma$ is computed by
\begin{align}\label{robust_elm_opt_new_4}
    &\min \gamma \nonumber
    \\
    \mathrm{s.t.}~&\begin{bmatrix}
        \gamma - \Omega_{1,1}  & 0 & \Omega_{1,3}
       \\
       * & \Omega_{2,2} & Z^{\top}
       \\
       * & * & I
    \end{bmatrix}\succeq 0 
\end{align}
in which $\Omega_{1,1} =\sum_{i=1}^{N}\sum_{j=1}^{n^{(1)}}\lambda_{i,j}$, $\Omega_{1,3}=\mathrm{vec}^{\top}(W^{(2)}H_0 - Y)$, $\Omega_{2,2} = \sum_{i=1}^{N}\sum_{j=1}^{n^{(1)}}\lambda_{i,j} Q_{i,j}$ and $Z= [\mathrm{vec}(W^{(2)}H_{1,1} - Y),\cdots,\mathrm{vec}(W^{(2)}H_{N,n^{(1)}} - Y)]$.
 
\end{proposition}
\begin{proof}
In order to develop a tractable algorithm to solve (\ref{robust_elm_opt}), we propose the following equivalent representation for interval matrix $H(\mathcal{U})$ in (\ref{H_interval}), i.e., 
\begin{align}\label{H_new_representation}
    H(\mathcal{U}) = H_0 + \sum_{i=1}^{N}\sum_{j = 1}^{n^{(1)}}\tau_{i,j}H_{i,j}
\end{align}
where $\tau_{i,j} \in [-1,1]$ and $H_0$, $H_{i,j}$ are defined by
\begin{align}\label{H0}
    H_0 &= \begin{bmatrix} 
   \frac{\underline{h}_{1,1}+\overline{h}_{1,1}}{2} & \cdots & \frac{\underline{h}_{1,n^{(1)}}+\overline{h}_{1,n^{(1)}}}{2}
   \\
    \vdots & \ddots & \vdots \\ 
    \frac{\underline{h}_{N,1}+\overline{h}_{N,1}}{2} & \cdots & \frac{\underline{h}_{N,n^{(1)}}+\overline{h}_{N,n^{(1)}}}{2}  
   \end{bmatrix}^{\top} 
   \\ \label{H_ij}
        H_{i,j} &=\begin{bmatrix}
    \mathbf{0}_{(i-1)\times(j-1)} &  \mathbf{0}_{(i-1)\times 1} &  \mathbf{0}_{(i-1)\times n^{(1)}-j}
    \\
     \mathbf{0}_{1 \times (j-1)} & \frac{\overline{h}_{i,j}-\underline{h}_{i,j}}{2}  &  \mathbf{0}_{1 \times (n^{(1)}-j)}
    \\
     \mathbf{0}_{(N-i)\times(j-1)} &  \mathbf{0}_{(N-i)\times 1} &  \mathbf{0}_{(N-i)\times n^{(1)}-j}
         \end{bmatrix} ^{\top} .
\end{align}

Based on (\ref{H_new_representation})--(\ref{H_ij}), $\left\|\mathrm{vec}(W^{(2)}H(\mathcal{U}) - Y\right)\|_2$ can be expressed as
\begin{align}
&\left\|\mathrm{vec}(W^{(2)}H(\mathcal{U}) - Y)\right\|_2  \nonumber
\\
=& \left\|\mathrm{vec}\left(W^{(2)}\left(H_0 + \sum_{i=1}^{N}\sum_{j = 1}^{n^{(1)}}\tau_{i,j}H_{i,j}\right) - Y\right)\right\|_2 \nonumber
\\
=& \xi^{\top} \Theta \xi \label{xi}
\end{align}
where $\xi = [1,\tau]^{\top}$, $\tau = [\tau_{1,1},\cdots,\tau_{N,n^{(1)}}]$ and 
\begin{align} \label{Theta}
    \Theta = \begin{bmatrix}
       \Theta_{1,1} & \Theta_{1,2}
       \\
       * & \Theta_{2,2}
    \end{bmatrix}
\end{align}
in which $\Theta_{1,1} = \mathrm{vec}^{\top}(W^{(2)}H_0 - Y)\mathrm{vec}(W^{(2)}H_0 - Y)$, $\Theta_{1,2} = \mathrm{vec}^{\top}(W^{(2)}H_0 - Y)Z$ with $Z = [\mathrm{vec}(W^{(2)}H_{1,1} - Y),\cdots,\mathrm{vec}(W^{(2)}H_{N,n^{(1)}} - Y)]$, and $\Theta_{2,2} = Z^{\top}Z$.

Therefore, the robust optimization problem (\ref{robust_elm_opt}) can be rewritten to 
\begin{align}\label{robust_elm_opt_new}
    &\min_{ W^{(2)}} \xi^{\top} \Theta \xi \nonumber
    \\
    \mathrm{s.t.}~&\left\|\tau\right\|_{\infty} \le 1
\end{align}
where $\Theta$ and $\xi$ are defined by (\ref{xi}) and (\ref{Theta}). 

Moreover, letting $Q_{i,j} = \mathrm{diag}\{\mathbf{0}_{1,i \times j -1},1,\mathbf{0}_{N \times n^{(1)} -i \times j}\}$, we can see that $\tau^{\top}Q_{i,j}\tau \le 1$ will deduce  $\left\|\tau\right\|_{\infty} \le 1$. Furthermore, $\tau^{\top}Q_{i,j}\tau \le 1$ equals to 
\begin{align}
    \xi^{\top}\begin{bmatrix}
       1 & 0
       \\
       * & -Q_{i,j}
    \end{bmatrix}\xi \ge 0 
\end{align}
Thus, we can formulate the following optimization problem \begin{align}\label{robust_elm_opt_new_2}
    &\min_{ W^{(2)}} \xi^{\top} \Theta \xi \nonumber
    \\
    \mathrm{s.t.}~&\xi^{\top}\begin{bmatrix}
       1 & 0
       \\
       * & -Q_{i,j}
    \end{bmatrix}\xi \ge 0
\end{align}
for all $i = 1,\ldots,N$, $j = 1,\ldots, n^{(1)}$. It is noted that the solution of (\ref{robust_elm_opt_new_2}) also satisfies optimization problem (\ref{robust_elm_opt_new}).

Using S-procedure and letting $\gamma = \xi^{\top} \Theta \xi$, we can formulate an optimization problem with $\lambda_{i,j} >0$ as follows
\begin{align}\label{robust_elm_opt_new_3}
    &\min \gamma \nonumber
    \\
    \mathrm{s.t.}~&\begin{bmatrix}
       \gamma - \Theta_{1,1} & -\Theta_{1,2}
       \\
       * & -\Theta_{2,2}
    \end{bmatrix}-\sum_{i=1}^{N}\sum_{j=1}^{n^{(1)}}\lambda_{i,j}\begin{bmatrix}
       1 & 0
       \\
       * & -Q_{i,j}
    \end{bmatrix}\succeq 0 
\end{align}
which ensures (\ref{robust_elm_opt_new_2}) holds.

Based on Schur Complement formula, it is equivalent to 
\begin{align}
    &\min \gamma \nonumber
    \\
    \mathrm{s.t.}~&\begin{bmatrix}
        \gamma - \Omega_{1,1}  & 0 & \Omega_{1,3}
       \\
       * & \Omega_{2,2} & Z^{\top}
       \\
       * & * & I
    \end{bmatrix}\succeq 0 
\end{align}
where $\Omega_{1,1} =\sum_{i=1}^{N}\sum_{j=1}^{n^{(1)}}\lambda_{i,j}$, $\Omega_{1,3}=\mathrm{vec}^{\top}(W^{(2)}H_0 - Y)$ and $\Omega_{2,2} = \sum_{i=1}^{N}\sum_{j=1}^{n^{(1)}}\lambda_{i,j} Q_{i,j}$. Therefore, we have that the optimized $\gamma \ge \xi^{\top}\Theta \xi$, which implies that $\gamma \ge \left\|\mathrm{vec}(W^{(2)}H(\mathcal{U}) - Y\right)\|_2$. The proof is complete. 
\end{proof}
\begin{remark}
Proposition \ref{proposition2} suggests that robust optimization problem (\ref{robust_elm_opt}) can be formulated in the form of Semi-Definite Programming (SDP) problem, so that the weights $W^{(2)}$ of the output layer can be efficiently solved with the help of existing SDP tools. By solving the optimization problem (\ref{robust_elm_opt_new_4}), the obtained weights $W^{(2)}$ are designed to make the approximation error $\gamma$ between disturbed input data set $\mathcal{U}$ and target set $Y$ as small as possible, which implies a robust training performance of shallow neural network (\ref{shallow}).    
\end{remark}
\begin{remark} \label{remark5}
Since the robust training process considers perturbations imposed on the input data set and optimizes weights to minimize the approximation error, the neural network tends to be able to tolerate noises better than those trained by the traditional training process. On the other hand, also due to the consideration of perturbations which are purposefully crafted in the input data set and is a player that is always playing against weights in robust optimization training, the approximation error would increase compared with traditional training. This is the trade-off between robustness and accuracy in neural network training, and it will be illustrated in a training example later.  
\end{remark}

In summary, the robust optimization training algorithm for shallow neural networks is presented in Algorithm \ref{alg}, which consists of three major components.

\begin{algorithm}[t!] 
\SetAlgoLined
\SetKwInOut{Input}{Input}
\SetKwInOut{Output}{Output}
\Input{Input data set $U$, output data set $Y$.}
\Output{Weights and biases $W^{k}$, $b^{(k)}$, $k=1,2$ for Shallow neural network  ${\Phi}$.}
\tcc{Initialization}
Generate perturbed input data interval set $\mathcal{U}$;
\\Randomly assign weights $W^{(1)}$ and biases $b^{(1)}$, and $b^{(2)} = 0$;
\\
\tcc{Reachable Set Computation}
Compute the  hidden layer output set $H(\mathcal{U})$ using (\ref{propostion1_1})--(\ref{propostion1_4});
\\
\tcc{Robust optimization}
Solve SDP problem (\ref{robust_elm_opt_new_4}) to obtain output layer weights $W^{(2)}$.
\caption{Robust Optimization Training of Shallow Neural Networks} \label{alg}
\end{algorithm}

\noindent\textbf{Initialization:} Since we employ ELM to train shallow neural networks, the weights $W^{(1)}$ and biases $b^{(1)}$ of the hidden layer are randomly assigned. In addition, the biases of output layer is set to $b^{(2)} = 0$. According to \cite{huang2006extreme}, $W^{(1)}$, $b^{(1)}$ and $b^{(2)}$ will remain unchanged in the rest of training process. 

\noindent \textbf{Reachable Set Computation:} The reachability analysis comes into play for the computation of $H(\mathcal{U})$, i.e., the reachable set of hidden layer. The computation is carried out based on (\ref{propostion1_1})--(\ref{propostion1_4}) in Proposition \ref{proposition1}.

\noindent\textbf{Robust Optimization:} This is the key step to achieve robustness in training shallow neural networks. Based on Proposition \ref{proposition2}, the robust optimization training process is converted to an SDP problem in the form of (\ref{robust_elm_opt_new_4}), which can be solved by various SDP tools. 

\begin{remark}
As shown in SDP problem (\ref{robust_elm_opt_new_4}), the computational cost of Algorithm \ref{alg} heavily depends on the number of decision variables $\lambda_{i,j}$ which is $Nn^{(1)}$. The value of $Nn^{(1)}$ is normally dominated by the number of input data $N$ which usually is a large number as sufficient input data is normally required for desired training performance. To reduce the computational cost in practice, we can modify $\lambda_{i,j}$ such as particularly letting $\tau_{1,j}=\cdots=\tau_{N,j}$ to relax the computational burden caused by a large number of input data. However, the price to pay here is that the result is a sub-optimal solution instead of the optimal solution to (\ref{robust_elm_opt_new_4}).
\end{remark}



\section{Evaluation}
\begin{figure}
\centering
	\includegraphics[width=3.5cm]{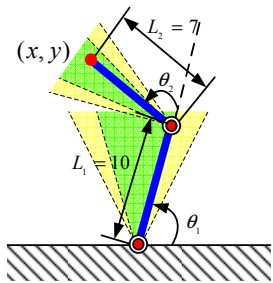}
	\caption{Robotic arm with two joints. The normal working zone of $(\theta_1,\theta_2)$ is colored in green $\theta_1,\theta_2 \in [\frac{5\pi}{12},\frac{7\pi}{12}]$. The buffering zone is in yellow $\theta_1,\theta_2 \in [\frac{\pi}{3},\frac{5\pi}{12}] \cup [\frac{7\pi}{12},\frac{2\pi}{3}] $. The forbidden zone is $\theta_1,\theta_2 \in [0,\frac{\pi}{3}] \cup [\frac{2\pi}{3},2\pi] $.}
	\label{robot_arm} 
\end{figure}

In this section, a \emph{learning forward kinematics} of a robotic arm model with two joints proposed in \cite{xiang2018output} is used to evaluate our developed robust optimization training method. The robotic arm model is shown in Fig. \ref{robot_arm}. 

The learning task is using a feedforward neural network to predict the position $(x,y)$ of the end with knowing the joint angles $(\theta_1,\theta_2)$. The input space $[0,2\pi]\times [0,2\pi]$ for $(\theta_1,\theta_2)$  is classified into three zones for its operations: Normal working zone $\theta_1,\theta_2 \in [\frac{5\pi}{12},\frac{7\pi}{12}]$, buffering zone $\theta_1,\theta_2 \in [\frac{\pi}{3},\frac{5\pi}{12}] \cup [\frac{7\pi}{12},\frac{2\pi}{3}] $ and forbidden zone $\theta_1,\theta_2 \in [0,\frac{\pi}{3}] \cup [\frac{2\pi}{3},2\pi]$. The detailed formulation for this robotic arm model and neural network training can be found in \cite{xiang2018output}. 

\begin{figure}
\centering
	\includegraphics[width=8.5cm]{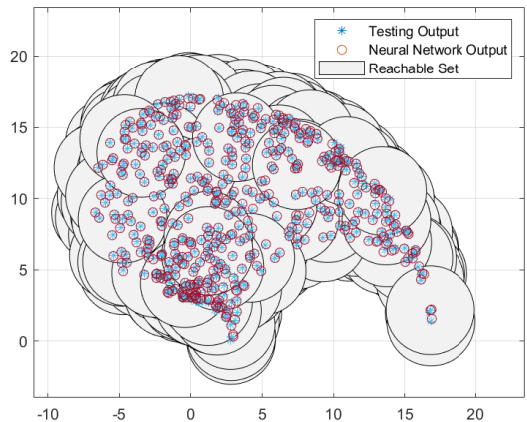}
	\caption{The reachable sets of the neural network trained by  traditional ELM  method for robot arm model. Maximal radius of output sets subject to disturbed input data set is  $3.106$.}\label{elm_fig}
\end{figure}

\begin{figure}
\centering
	\includegraphics[width=8.5cm]{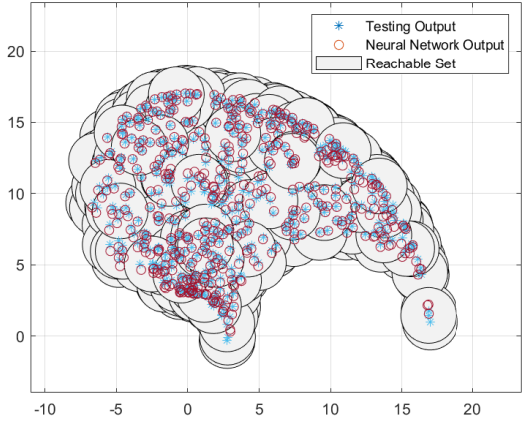}
	\caption{The reachable sets of the neural network trained by robust optimization learning method for robot arm model. The maximal radius of output sets subject to disturbed input data set is  $1.9494$ which means the robustness has been improved. On the other hand, the deviation of neural network outputs from output data is increased by our observation, which means the training accuracy has been decreased. Therefore, the trade-off between robustness and training accuracy exists.  }\label{roelm_fig}
\end{figure}

To show the advantage of robust learning, we first train a shallow neural network using the traditional ELM method. Then, assuming the injected disturbances are $\delta_i = 0.01$. By using Lemma \ref{lem_outputset} and choosing the maximal deviation of outputs as the radius for all testing output data, the output reachable set for all perturbed inputs are shown in Fig. \ref{elm_fig}. Moreover, we train a shallow neural network using Algorithm \ref{alg}, i.e., the robust optimization training method. The output sets for perturbed inputs are shown in Fig. \ref{roelm_fig}. It can be explicitly observed that the neural network trained by the robust optimization method has tighter reachable sets which means the neural network is less sensitive to disturbances. Therefore, we can conclude that the neural network trained by the robust optimization method is more robust to noises injected in input data. On the other hand, comparing Figs. \ref{elm_fig} and \ref{roelm_fig}, the deviation of neural network output from output data is increased by observation, i.e, the training accuracy is sacrificed for improving robustness.

Furthermore, the trade-off between robustness and accuracy mentioned in Remark \ref{remark5} are elucidated in Table \ref{tab1}. It can be seen that robust learning provides a better tolerance in input data noises but yields less accuracy than the traditional learning process, i.e., a larger Mean Square Error (MSE).  

\begin{table}
	\centering
	\caption{Radius of Output Reachable Sets and Mean Square Error}
	\label{tab1}
	\begin{tabular}{|ccc|}
		\hline 
\textbf{Method} & \textbf{Radius}	& \textbf{MSE}\\
		\hline\hline
Traditional ELM  &$3.1060$  &$0.0076$  \\
\hline
Algorithm \ref{alg}   &$1.9494$ &$0.0643$\\
\hline
	\end{tabular}
\end{table}

\section{Conclusions}
A robust optimization learning framework is proposed in this paper for shallow neural networks. First, the input set data are generalized to interval sets to characterize injected noises. Then
based on the layer-by-layer reachability analysis for neural networks, the output sets of the hidden layer are computed, which play a critical role in the robust optimization training process. The robust training problem is formulated in terms of robust least-squares problems, which can be then converted to an SDP problem. The trade-off between robustness and training accuracy is observed in the proposed framework. A robot arm modeling example is provided to evaluate our method.

\bibliographystyle{IEEEtran}
\bibliography{reference}

\begin{thebibliography}{10}
\providecommand{\url}[1]{#1}
\csname url@samestyle\endcsname
\providecommand{\newblock}{\relax}
\providecommand{\bibinfo}[2]{#2}
\providecommand{\BIBentrySTDinterwordspacing}{\spaceskip=0pt\relax}
\providecommand{\BIBentryALTinterwordstretchfactor}{4}
\providecommand{\BIBentryALTinterwordspacing}{\spaceskip=\fontdimen2\font plus
\BIBentryALTinterwordstretchfactor\fontdimen3\font minus
  \fontdimen4\font\relax}
\providecommand{\BIBforeignlanguage}[2]{{%
\expandafter\ifx\csname l@#1\endcsname\relax
\typeout{** WARNING: IEEEtran.bst: No hyphenation pattern has been}%
\typeout{** loaded for the language `#1'. Using the pattern for}%
\typeout{** the default language instead.}%
\else
\language=\csname l@#1\endcsname
\fi
#2}}
\providecommand{\BIBdecl}{\relax}
\BIBdecl

\bibitem{hunt1992neural}
K.~J. Hunt, D.~Sbarbaro, R.~{\.Z}bikowski, and P.~J. Gawthrop, ``Neural
  networks for control systems—a survey,'' \emph{Automatica}, vol.~28, no.~6,
  pp. 1083--1112, 1992.

\bibitem{ge1999adaptive}
S.~S. Ge, C.~C. Hang, and T.~Zhang, ``Adaptive neural network control of
  nonlinear systems by state and output feedback,'' \emph{IEEE Transactions on
  Systems, Man, and Cybernetics, Part B (Cybernetics)}, vol.~29, no.~6, pp.
  818--828, 1999.

\bibitem{wang2015combined}
T.~Wang, H.~Gao, and J.~Qiu, ``A combined adaptive neural network and nonlinear
  model predictive control for multirate networked industrial process
  control,'' \emph{IEEE Transactions on Neural Networks and Learning Systems},
  vol.~27, no.~2, pp. 416--425, 2015.

\bibitem{wu2014exponential}
Z.-G. Wu, P.~Shi, H.~Su, and J.~Chu, ``Exponential stabilization for
  sampled-data neural-network-based control systems,'' \emph{IEEE Transactions
  on Neural Networks and Learning Systems}, vol.~25, no.~12, pp. 2180--2190,
  2014.

\bibitem{tian2018deeptest}
Y.~Tian, K.~Pei, S.~Jana, and B.~Ray, ``Deeptest: Automated testing of
  deep-neural-network-driven autonomous cars,'' in \emph{Proceedings of the
  40th International Conference on Software Engineering}, 2018, pp. 303--314.

\bibitem{neumann2013neural}
K.~Neumann, A.~Lemme, and J.~J. Steil, ``Neural learning of stable dynamical
  systems based on data-driven lyapunov candidates,'' in \emph{2013 IEEE/RSJ
  International Conference on Intelligent Robots and Systems}.\hskip 1em plus
  0.5em minus 0.4em\relax IEEE, 2013, pp. 1216--1222.

\bibitem{he2020learnability}
H.~He, M.~Chen, G.~Xu, Z.~Zhu, and Z.~Zhu, ``Learnability and robustness of
  shallow neural networks learned with a performance-driven bp and a variant
  pso for edge decision-making,'' \emph{arXiv preprint arXiv:2008.06135}, 2020.

\bibitem{madry2017towards}
A.~Madry, A.~Makelov, L.~Schmidt, D.~Tsipras, and A.~Vladu, ``Towards deep
  learning models resistant to adversarial attacks,'' \emph{arXiv preprint
  arXiv:1706.06083}, 2017.

\bibitem{madaan2020adversarial}
D.~Madaan, J.~Shin, and S.~J. Hwang, ``Adversarial neural pruning with latent
  vulnerability suppression,'' in \emph{International Conference on Machine
  Learning}.\hskip 1em plus 0.5em minus 0.4em\relax PMLR, 2020, pp. 6575--6585.

\bibitem{wong2018provable}
E.~Wong and Z.~Kolter, ``Provable defenses against adversarial examples via the
  convex outer adversarial polytope,'' in \emph{International Conference on
  Machine Learning}.\hskip 1em plus 0.5em minus 0.4em\relax PMLR, 2018, pp.
  5286--5295.

\bibitem{zhang2019theoretically}
H.~Zhang, Y.~Yu, J.~Jiao, E.~Xing, L.~El~Ghaoui, and M.~Jordan, ``Theoretically
  principled trade-off between robustness and accuracy,'' in
  \emph{International Conference on Machine Learning}.\hskip 1em plus 0.5em
  minus 0.4em\relax PMLR, 2019, pp. 7472--7482.

\bibitem{yao2020lightweight}
D.~Yao, H.~Liu, J.~Yang, and X.~Li, ``A lightweight neural network with strong
  robustness for bearing fault diagnosis,'' \emph{Measurement}, vol. 159, p.
  107756, 2020.

\bibitem{xiang2017reachable}
W.~Xiang, H.-D. Tran, and T.~T. Johnson, ``Reachable set computation and safety
  verification for neural networks with \textsc{r}e\textsc{lu} activations,''
  \emph{arXiv preprint arXiv:1712.08163}, 2017.

\bibitem{xiang2018output}
------, ``Output reachable set estimation and verification for multilayer
  neural networks,'' \emph{IEEE Transactions on Neural Networks and Learning
  Systems}, vol.~29, no.~11, pp. 5777--5783, 2018.

\bibitem{xiang2018reachable}
W.~Xiang, H.-D. Tran, J.~A. Rosenfeld, and T.~T. Johnson, ``Reachable set
  estimation and safety verification for piecewise linear systems with neural
  network controllers,'' in \emph{2018 Annual American Control Conference
  (ACC)}.\hskip 1em plus 0.5em minus 0.4em\relax IEEE, 2018, pp. 1574--1579.

\bibitem{tran2019star}
H.-D. Tran, D.~M. Lopez, P.~Musau, X.~Yang, L.~V. Nguyen, W.~Xiang, and T.~T.
  Johnson, ``Star-based reachability analysis of deep neural networks,'' in
  \emph{International Symposium on Formal Methods}.\hskip 1em plus 0.5em minus
  0.4em\relax Springer, 2019, pp. 670--686.

\bibitem{xiang2018reachability}
W.~Xiang and T.~T. Johnson, ``Reachability analysis and safety verification for
  neural network control systems,'' \emph{arXiv preprint arXiv:1805.09944},
  2018.

\bibitem{huang2006extreme}
G.-B. Huang, Q.-Y. Zhu, and C.-K. Siew, ``Extreme learning machine: theory and
  applications,'' \emph{Neurocomputing}, vol.~70, no. 1-3, pp. 489--501, 2006.

\bibitem{chen2017broad}
C.~P. Chen and Z.~Liu, ``Broad learning system: An effective and efficient
  incremental learning system without the need for deep architecture,''
  \emph{IEEE Transactions on Neural Networks and Learning Systems}, vol.~29,
  no.~1, pp. 10--24, 2017.

\bibitem{xiang2020reachable}
W.~Xiang, H.-D. Tran, X.~Yang, and T.~T. Johnson, ``Reachable set estimation
  for neural network control systems: a simulation-guided approach,''
  \emph{IEEE Transactions on Neural Networks and Learning Systems}, 2020, DOI:
  10.1109/TNNLS.2020.2991090.

\end{thebibliography}
\appendix

\subsection{Proof of Lemma \ref{lem_outputset}}
We shall focus on $\mathcal{X}_k \subseteq [\underline{x}^{(k)},\overline{x}^{(k)}],\forall k=1,2$ . Considering the $k$th layer, the following inequalities 
\begin{align*}
   w_{i,j}^{(k)} x^{(k-1)}_j \geq \underline{p}_{i,j}
  \\
  w_{i,j}^{(k)} x^{(k-1)}_j \le \overline{p}_{i,j}
\end{align*}
hold for any input of this layer $x^{(k-1)} \in [\underline{x}^{(k-1)},\overline{x}^{(k-1)}]$.

Using the monotonic property (\ref{mono})  of activation function $\phi_k$ in Assumption \ref{assumption_mono}, it leads to
\begin{align}\label{input interval}
        x_{i}^{(k+1)} \ge \phi_k(\sum\nolimits_{j=1}^{n_{k}}\underline{p}^{(k)}_{i,j}+b_i)
    \\
    x_{i}^{(k+1)} \le \phi_k(\sum\nolimits_{j=1}^{n_{k}}\overline{p}^{(k)}_{i,j}+b_i)
\end{align}
for any input $x^{(k-1)} \in [\underline{x}^{(k-1)},\overline{x}^{(k-1)}]$.  Thus, by iterating the above process from $k = 1,2$, given any $x^{(0)} \in [\underline{x}^{(0)},\overline{x}^{(0)}]$, the output of the neural network (\ref{nn}) satisfies  
\begin{align}\label{rec_form}
    x^{(k)} \in [\underline{x}^{(k)},\overline{x}^{(k)}],~k=1,2
\end{align}
where $[\underline{x}^{(k)},\overline{x}^{(k)}]$ covers the outputs of reachable set computation (\ref{reachable set computation}) for the $k$th layer, respectively. Therefore, for any input set $\mathcal{X}_0\subseteq [\underline{x}^{(0)},\overline{x}^{(0)}]$, the output set $\mathcal{X}_k \subseteq [\underline{x}^{(k)},\overline{x}^{(k)}],\forall k = 1,2$. The proof is complete.

\end{document}